\documentclass[letterpaper]{article}
\usepackage{epsfig,fullpage,amsmath}
\usepackage{aaai}
\usepackage{times}
\usepackage{helvet}
\usepackage{courier}

\begin{document}

\newcommand{\remove}[1]{}
\newcommand{\dima}[1]{{{#1}}}
\newcommand{\zchaff}{{\tt zChaff}}
\newcommand{\oksolver}{{\tt OKsolver}}
\newcommand{\satz}{{\tt Satz}}
\newcommand{\walksat}{{\tt WalkSAT}}
\newcommand{\survey}{{\tt SP}}
\newcommand{\uc}{{\tt UC}}

\newcommand{\whp}{w.h.p.}
\newcommand{\ie}{i.e.,\ }
\newcommand{\ex}{{\bf E}}
\def\eg{e.g.\ }

\newcommand{\ds}{{\rm d}s}
\newcommand{\dc}{{\rm d}c}
\newcommand{\dr}{{\rm d}r}
\newcommand{\dx}{{\rm d}x}

\title{Generating Hard Satisfiable Formulas by Hiding Solutions Deceptively}

\author{Haixia Jia 
\and Cristopher Moore \and Doug Strain \\ Computer Science Department\\
University of New Mexico \\ {\tt \{hjia,moore\}@cs.unm.edu, doug.strain@gmail.com}}

\date{}
\maketitle

\begin{abstract}
To test incomplete search algorithms for constraint satisfaction problems 
such as 3-SAT, we need a source of hard, but satisfiable, benchmark instances.  
A simple way to do this is to choose a random truth assignment $A$, 
and then choose clauses randomly from among those satisfied by $A$. 
However, this method tends to produce easy problems, since 
the majority of literals point toward the ``hidden'' assignment $A$.  
Last year,~\cite{2hidden} proposed a problem generator that cancels 
this effect by hiding both $A$ and its complement $\overline{A}$.  
While the resulting formulas appear to be just as hard for DPLL algorithms 
as random 3-SAT formulas with no hidden assignment, they can be solved
by \walksat\ in only polynomial time.  

Here we propose a new method to cancel the attraction to $A$, 
by choosing a clause with $t > 0$ literals satisfied by $A$ with probability 
proportional to $q^t$ for some $q < 1$.  
By varying $q$, we can generate formulas whose variables have no bias, 
i.e., which are equally likely to be true or false; we can even cause the formula 
to ``deceptively'' point away from $A$.  We present theoretical and experimental 
results suggesting that these formulas are exponentially hard both for DPLL algorithms 
and for incomplete algorithms such as \walksat.
\end{abstract}

\section{Introduction}

To evaluate search algorithms for constraint satisfaction problems, 
we need good sources of benchmark instances.  
Real-world problems are the best benchmarks by definition, 
but each such problem has structures specific to its application domain; 
in addition, if we wish to gather good data on how the running times of 
our algorithms scale, we need entire families of benchmarks with 
varying size and density.  

One way to fill this need is to generate {\em random} instances.  For instance, 
for 3-SAT we can generate instances with $n$ variables and $m$ clauses 
by choosing each clause uniformly from among the $8 {n \choose 3}$ possibilities.  
We can then vary these formulas according to their size and their density $r=m/n$.  
While such formulas lack much of the structure of real-world instances, 
they have been instrumental in the development and study of new search methods 
such as simulated annealing~\cite{Johnson}, 
the breakout procedure~\cite{Morris}, 
\walksat~\cite{Selman96}, 
and Survey Propagation~\cite{mz}.

However, if we wish to test {\em incomplete} algorithms such as 
\walksat\ and Survey Propagation (\survey), we need a source of hard, 
but satisfiable problems.  In contrast, above a critical density 
$r \approx 4.27$, the random formulas defined above are almost certainly unsatisfiable.  
Random formulas at this threshold appear to be quite hard for 
complete solvers~\cite{cheeseman,hh,msl}; but for precisely this reason, 
it is not feasible to generate large problems at the threshold and then 
filter out the unsatisfiable ones.  While other classes of satisfiable CSPs
have been proposed, such as the quasigroup completion 
problem~\cite{ssw,quasigroup1,quasigroup2}, we would like to have
problems generators that are ``native'' to 3-SAT.

A natural way to generate random satisfiable 3-SAT formulas is to 
choose a random truth assignment $A \in \{0,1\}^n$, and then choose $m$ clauses 
uniformly and independently from among the $7 {n \choose 3}$ clauses satisfied 
by $A$.  The problem with this is that simply rejecting clauses that conflict with $A$ 
causes an unbalanced distribution of literals; in particular, 
on average a literal will agree with its value in the hidden assignment $4/7$ of the time.  
Thus, especially when there are many clauses, a simple majority heuristic or local search will 
quickly find $A$.  More sophisticated versions of this ``hidden assignment''
scheme~\cite{Asahiro,VanGelder} improve matters somewhat but still
lead to biased samples.  Thus the question is how to avoid this ``attraction'' to the 
hidden assignment,

One approach~\cite{2hidden}
is to choose clauses uniformly from among those 
that are satisfied by both $A$ and its complement $\overline{A}$.  
This is inspired by recent work on random $k$-SAT and Not-All-Equal SAT~\cite{achmooreksat},  
in which symmetry with respect to complementation reduces the variance of
the number of solutions; 
the idea is that $A$ and $\overline{A}$ 
cancel each others' attractions out, making either one hard to find.  Indeed, the resulting formulas appear 
to take DPLL solvers exponential time and, in general, to be just as hard as random 3-SAT formulas
with no hidden assignment.  On the other hand, \walksat\ solves these formulas in
polynomial time, since after a few variables are set in a way that agrees with 
one of the hidden assignments, 
neighboring variables develop correlations consistent with these~\cite{Barthel}.

In this paper, we pursue an alternate approach, inspired by~\cite{achperes}, 
who reweighted the satisfying assignments in a natural way.   
We hide just one assignment, but we bias the distribution of clauses as follows:  
for each clause, we choose a random $3$-tuple 
(or more generally, a $k$-tuple) of variables, and construct a clause 
with $t > 0$ literals satisfied by $A$ 
with probability proportional to $q^t$ for some constant $q < 1$.  
(Note that the naive formulas discussed above amount to the case $q=1$.)
This penalizes the clauses which are ``more satisfied'' by $A$, 
and reduces the extent to which variable occurrences are more likely to agree with $A$.  
As we will see below, by choosing $q$ appropriately we can rebalance the distribution 
of literals, so that each variable is as likely to appear positively as often as negatively 
and no longer points toward its value in $A$.  By reducing $q$ further, 
we can even make it more likely that a variable occurrence {\em disagrees} with $A$, 
so that the formula becomes ``deceptive'' and points away from the hidden assignment.

\remove{
\begin{enumerate}
\item Predefine a constant $q \leq 1$ and generate a random truth assignment $A$;
\item Randomly pick $k$ variables from $n$ variables, 
form a certain type of clause which is satisfied by $A$ with probability
$\frac{q^p}{(1+q)^k-1}$ where $p$ is the number of literals that agree with
the truth assignment.
\item Repeat step 2 $rn$ times to generate a formula with $n$ variables and
$rn$ clauses.
\end{enumerate}
}

We call these formulas ``$q$-hidden,'' to distinguish them from the naive ``1-hidden'' 
formulas discussed above, the ``2-hidden'' formulas studied in~\cite{2hidden}, and 
the ``0-hidden'' formulas consisting of random 3-SAT formulas with no hidden assignment.  
Like these other families, our $q$-hidden formulas are 
readily amenable to all the mathematical tools that have been
developed for studying random $k$-SAT formulas, including moment calculations
and the method of differential equations.  Below we calculate the expected density
of satisfying assignments as a function of their distance from $A$, and analyze the 
behavior of the Unit Clause (\uc) algorithm on $q$-hidden formulas. 
We then present experiments on several complete and incomplete solvers.  
We find that our $q$-hidden formulas are just as hard for DPLL algorithms as 0-hidden 
formulas, and are much harder than naive 1-hidden formulas.  In addition, we find that
local search algorithms like \walksat\ find our formulas much harder than any of these
other families, taking exponential as opposed to polynomial time.  Moreover, the 
running time of \walksat\ increases sharply as our formulas become more deceptive.

\section{The expected density of solutions}
\label{sec:structure}

For $\alpha \in [0,1]$, let $X_\alpha$ be the number of satisfying 
truth assignments in a random $q$-hidden $k$-SAT formula that agree on a fraction
$\alpha$ of the variables with the hidden assignment $A$; that is, 
their Hamming distance from $A$ is $(1-\alpha)n$.  We wish to calculate
the expectation $\ex[X_\alpha]$.  

By symmetry, we can take $A$ to be the all-true assignment.  In that case, 
a clause with $t > 0$ positive literals is chosen with probability 
\[ \frac{q^t}{(1+q)^k-1} \enspace . \]
Let $B$ be a truth assignment where $\alpha n$ of the variables are true 
and $(1-\alpha)n$ are false.  
Then, analogous to~\cite{2hidden}, we use 
linearity of expectation, independence between clauses, 
the selection of the literals in each clause with replacement, and 
Stirling's approximation for the factorial to obtain 
(where $\sim$ suppresses terms polynomial in $n$): 

\begin{align*}
\ex[X_\alpha]
&= {n \choose \alpha n} \Pr[\mbox{$B$ satisfies a random clause}]^m \\
&= {n \choose \alpha n} \left( 1 - \sum_{t=1}^k {k \choose t} 
  \frac{q^t (1-\alpha)^t \alpha^{k-t}}{(1+q)^k-1}  \right)^{\!m} \\
&\sim f_{k,r,q}(\alpha)^n
\end{align*}
where
\[ f(\alpha) = \frac{1}{\alpha^\alpha (1-\alpha)^{1-\alpha}} 
\left(1-\frac{(q(1-\alpha) + \alpha)^k -\alpha^k}
{(1+q)^k-1}\right)^{\!r}
\enspace . 
\]

\begin{figure}
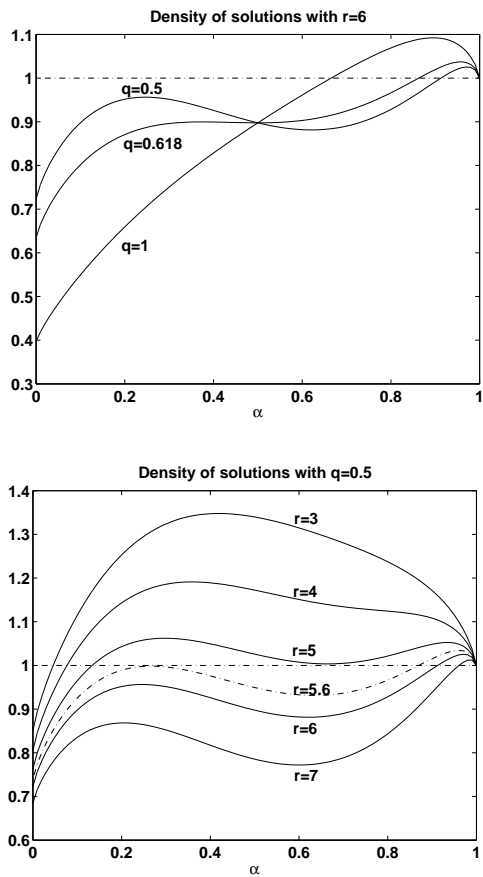

\begin{center}
\begin{minipage}{3in}
\begin{center}
        \centerline{\hbox{
        \psfig{figure=space3c.ps,width=2.5in}
        }}
\end{center}
\end{minipage}\    \
\begin{minipage}{3in}
\begin{center}
        \centerline{\hbox{
        \psfig{figure=space5.ps,width=2.5in}
        }}
\end{center}
\end{minipage} \ \
\end{center}
\caption{The $n$th root $f(\alpha)$ of the expected number of solutions which agree 
with the hidden assignment on a fraction $\alpha$ of the variables.  Here $k=3$.  
The upper part of the figure shows $f(\alpha)$ for $q=1$, $q=0.618$ and $q=0.5$ 
at $r=6$. The lower part shows $f(\alpha)$ for $q=0.5$ and varying $r$.  Note that at 
$r=5.6$, we have $f(\alpha) < 1$ for all $\alpha \le 1/2$.}
\label{fig:q6}
\end{figure}

Looking at Figure~\ref{fig:q6}, we see that the behavior of $f$ near $\alpha = 1/2$ 
changes dramatically as we vary $q$.  For $q=1$ (i.e., naive 1-hidden formulas), 
$f'(1/2)$ is positive, giving local search algorithms a ``push'' towards the hidden assignment.  
On the other hand, if $q$ is the positive root $q^*$ of
\[ (1-q)(1+q)^{k-1} - 1 = 0 \]
then $f'(1/2)=0$.  Analogous to~\cite{achperes}, this is also the value of $q$ 
at which literals are equally likely to agree or disagree with $A$.  Intuitively, then, 
if $q=q^*$ we would expect a local search algorithm starting from a 
random assignment---for which $\alpha$ is tightly concentrated around $1/2$---to have
no local information telling it in which direction the hidden assignment lies.  
We call these $q^*$-hidden formulas {\em balanced}; 
for $k=3$, $q^*$ is the golden ratio $(\sqrt{5}-1)/2=0.618...$ 

For smaller values of $q$ such as $q=0.5$ shown in Figure~\ref{fig:q6}, $f'(1/2)$ becomes negative, 
and we expect a local search algorithm starting at a random assignment to 
move {\em away} from $A$.  Indeed, $f(\alpha)$ has a local maximum at some 
$\alpha < 1/2$, and for small $r$ there are solutions with $\alpha < 1/2$.  
When $r$ is sufficiently large, however, $f(\alpha) < 1$ 
for all $\alpha < 1/2$, and the probability any of these ``alternate'' solutions exist
is exponentially small.  We conjecture that for each $q \le q^*$ there is a threshold $r_c(q)$ 
at which with high probability the only solutions are those close to $A$.  
Setting $\max \{f(\alpha) \mid \alpha \le 1/2\} = 1$ yields an upper bound on $r_c(q)$, which 
we show in Figure~\ref{fig:rc} below. For instance, $r_c(0.5) \le 5.6$ as shown 
in Figure~\ref{fig:q6}.

We call such formulas {\em deceptive}, 
since local search algorithms such as \walksat, DPLL algorithms such as \zchaff\ 
that use a majority heuristic in their splitting rule, and message-passing algorithms such as 
\survey\ will presumably search in the wrong direction, and take exponential 
time to cross the local minimum in $f(\alpha)$ to find the hidden assignment.  
Our experiments below appear to confirm this intuition.  In addition, all three types of 
algorithms appear to encounter the most difficulty at roughly the same density $r_c(q)$, 
where we conjecture the ``alternate'' solutions disappear.

\remove{
\begin{figure}
\begin{center}
\begin{minipage}{3in}
\begin{center}
        \centerline{\hbox{
        \psfig{figure=zchaffr.ps,width=2.5in}
        }}
\end{center}
\end{minipage}\    \
\begin{minipage}{3in}
\begin{center}
        \centerline{\hbox{
        \psfig{figure=ucthresh.ps,width=2.5in}
        }}
\end{center}
\end{minipage}
\end{center}
\caption{The upper part of the figure shows the minimum density $r_c$ at which the only solutions
exist at $\alpha >1/2$ derived from our calculation of solution
space and $r^*$ derived from \zchaff, \walksat\ and \survey as a function of $q$; 
the lower part of the figure shows
the density at which \uc\ begins to fail as a funtion of $q$.}
\label{fig:thresh}
\end{figure}
}

\section{Unit Clause heuristic and DPLL algorithms}
\label{sec:uc}

Unit Clause (\uc) is a linear-time heuristic which permanently 
sets one variable in each step as follows: if there are any unit clauses, 
satisfy them; otherwise, pick a random literal and satisfy it.  
For random 3-SAT formulas, \uc\ succeeds
with constant probability for $r < 8/3$,
and fails with high probability for $r > 8/3$~\cite{chaofranco}.
\uc\ can be thought as the first branch of a simple DPLL algorithm $S$, 
whose splitting rule takes a random unset variable and tries its truth values
in random order; thus \uc\ succeeds if $S$ succeeds without backtracking.
On the other hand,~\cite{monassontcs,monassonbook} showed that $S$'s 
expected running time is exponential in $n$ for any $r > 8/3$; see also~\cite{abm}, 
who used lower bounds on resolution complexity to show that $S$ takes 
exponential time with high probability if $r > 3.81$.  
In general, it appears that simple DPLL algorithms begin to take exponential time
at exactly the density where the corresponding linear-time heuristic fails.

In this section, we analyze the performance of 
\uc\ on our $q$-hidden formulas.  Specifically, we show that in the 
balanced case where $q=q^*$, \uc\ 
fails for $r > 8/3$ just as for 0-hidden formulas.    
Based on this, we conjecture that the running time of $S$, and other
simple DPLL algorithms, is exponentially large for our formulas
at the same density as for 0-hidden ones.  

As in~\cite{2hidden}, we 
analyze the behavior of \uc\ on arbitrary initial distributions of 3-SAT clauses 
using the method of differential equations. 
For simplicity we assume that $A$ is the all-true assignment. 
A {\em round} of \uc\ consists of a ``free step,'' 
in which we satisfy a random literal, and the 
ensuing chain of unit-clause propagations. 
For $0 \le i \le 3$ and $0 \le j \le i$, let $S_{i,j} = s_{i,j} n$ be
the number of clauses of length $i$ with $j$ positive literals and
$i-j$ negative ones, and $s_i = \sum_j s_{i,j}$.  
Let $X=xn$ be the
number of variables set so far, and let $m_T$ and $m_F$ be
the expected number of variables set true and false in a
round.  Then we can 
model the discrete stochastic process of the $S_{i,j}$ with the following
differential equations for the $s_{i,j}$:
\begin{eqnarray}
\frac{\ds_{3,j}}{\dx} & = & -\frac{3 s_{3,j}}{1-x}
\label{eq:rescaled} \\
\frac{\ds_{2,j}}{\dx} & = & -\frac{2 s_{2,j}}{1-x}
+ \frac{m_F (j+1) s_{3,j+1} + m_T (3-j) s_{3,j}}
 {(m_T + m_F)(1-x)} \nonumber
\end{eqnarray}
The unit clauses are governed by a two-type branching process, 
with transition matrix
\[ M = \frac{1}{1-x} \left( \begin{array}{cc}
s_{2,1} & 2 s_{2,0} \\
2 s_{2,2} & s_{2,1}
\end{array} \right) \enspace .
\]
As in~\cite{achmoorecoloring}, as long as the largest eigenvalue of $M$ is less than $1$, 
the branching process is subcritical, 
and summing over the round gives
\[ \left( \! \begin{array}{c} m_F \\ m_T \end{array} \! \right)
= \left( I - M \right)^{-1} \cdot \left( \! \begin{array}{c} 1/2 \\ 1/2 \end{array} \! \right) \enspace . \]
We then solve the equation~\eqref{eq:rescaled} with the initial conditions 
$s_{3,0} = 0$ and
\[ s_{3,j} = {3 \choose j} \frac{q^j}{(1+q)^3-1} \]
for $0 < j \le 3$.  In the balanced case $q = q^*$, 
we find that \uc\ succeeds on q-formulas
with constant probability if and only if $r < 8/3$, just as for 
0-hidden formulas.  The reason is that, as for 2-hidden formulas, 
the expected number of positive and negative literals are the same throughout 
the process.  This symmetry causes \uc\ to behave just as it would on random 3-SAT formulas
without a hidden assignment.  

We note that for $q < q^*$, \uc\ succeeds at slightly higher densities, 
at which it can find one of the ``alternate'' solutions with $\alpha < 1/2$.
At higher densities where these alternate solutions disappear, 
our experimental results below show that these ``deceptive'' formulas 
take DPLL algorithms exponential time, and for $r > r_c(q)$ they are  
harder than 0-hidden formulas of the same density.

\section{Experimental results}
\label{sec:exp}

\subsection{DPLL}

\begin{figure}[ht]
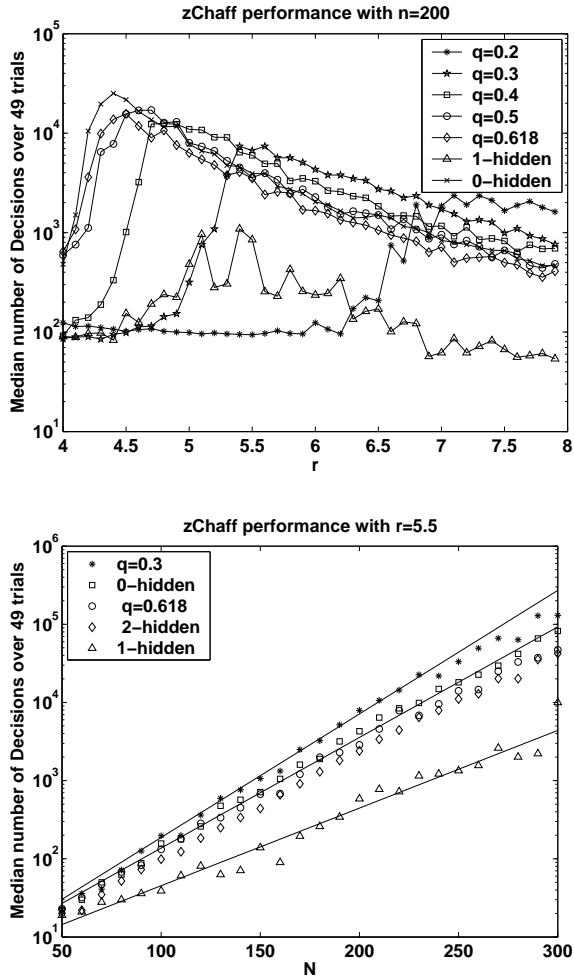

\begin{center}
\begin{minipage}{3in}
\begin{center}
        \centerline{\hbox{
        \psfig{figure=zchaffr.ps,width=3in}
        }}
\end{center}
\end{minipage}\ \
\begin{minipage}{3in}
\begin{center}
        \centerline{\hbox{
        \psfig{figure=zchaffn.ps,width=3in}
        }}
\end{center}
\end{minipage}
\end{center}
\vspace*{-.6cm}
\caption{The upper part of the figure shows \zchaff's median running time over 49 trials on 
0-hidden, 1-hidden and $q$-hidden formulas with $n=200$ and $r$ ranging from 4.0 to 8.0.  
The lower part shows the median running time with $r=5.5$ and $n$ ranging from 50 to 300.}
\label{fig:zchaff}
\end{figure}

In this section we discuss the behavior of DPLL solvers on our $q$-hidden formulas.  
We focus on \zchaff~\cite{zchaffsite}; our results from \oksolver~\cite{oksolver} are 
qualitatively similar.  Figure~\ref{fig:zchaff} shows \zchaff's median running time 
on 0-hidden, 1-hidden, and $q$-hidden 
formulas for various values of $q$.  We see the following phenomena:


Our $q$-hidden formulas with $q=q^*=0.618...$ are about as hard as 0-hidden ones, 
and peak in complexity near the satisfiability threshold. 
This is consistent with the picture given in the previous two sections: namely, 
that these ``balanced'' formulas make it impossible for 
algorithms to feel the attraction of the hidden assignment.  
In contrast, naive 1-hidden formulas are far easier, 
since the attraction to the hidden assignment is strong.

The $q$-hidden formulas with $q < q^*$ are the most interesting ones.  
The hardness of these formulas shows two phases: at low density they are relatively
easy, and their hardness peaks at a density $r_c(q)$.  Above $r_c(q)$ 
they take exponential time; as for 0-hidden formulas, although as $r$ increases further 
the coefficient of the exponential decreases as the clauses 
generate contradictions more quickly.

We believe that this peak $r_c(q)$ is the same threshold density defined earlier 
(see Figure~\ref{fig:rc} below)
above which the only solutions are those close to the hidden assignment.  
The situation seems to be the following: below $r_c(q)$, there are ``alternate'' 
solutions with $\alpha < 1/2$, and \zchaff\ is led to these by its splitting rule.  Above $r_c(q)$, 
these alternate solutions disappear, and \zchaff\ takes exponential time to find 
the vicinity of the hidden assignment, since the formula deceptively points in the other direction.  
Moreover, for a fixed $r$ above $r_c(q)$ these formulas become harder as $q$ decreases 
and they become more deceptive.  

To illustrate this further, the lower part of
Figure~\ref{fig:zchaff} shows \zchaff's median running time on 0-hidden formulas, 1-hidden formulas,
and $q$-hidden formulas for $q=q^*$ (balanced) and $q=0.3$ (deceptive).  
We also compare with the 2-hidden formulas of~\cite{2hidden}.  
We fix $r=5.5$, which appears to be above $r_c(q)$ for both these values of $q$.  
At this density, the 0-hidden, 2-hidden, and balanced $q$-hidden 
formulas are all comparable in difficulty, while 1-hidden formulas are much easier and 
the deceptive formulas appear to be somewhat harder.

\subsection{\survey}

Survey Propagation or \survey~\cite{mz} is a recently introduced incomplete solver 
based on insights from the replica method of statistical physics and 
a generalization of belief propagation.  We tested \survey\ on 0-hidden
formulas and $q$-hidden formulas for different values of $q$, using $n=10^4$ 
and varying $r$.  For 0-hidden formulas, \survey\ succeeds up to $r=4.25$, 
quite close to the satisfiability threshold.  For $q$-hidden formulas 
with $q=q^*$, \survey\ fails at $4.25$ just as it does for 0-hidden formulas, 
suggesting that it finds these formulas exactly as hard as 0-hidden ones 
even though they are guaranteed to be satisfiable.  For naive 1-hidden formulas, 
\survey\ succeeds at a significantly higher density, up to $r=5.6$.

Presumably the naive 1-hidden formulas are easier for \survey\ since the
``messages'' from clauses to variables, like the majority
heuristic, tend to push the algorithm towards the hidden assignment.
In the balanced case $q=q^*$, this attraction is 
successfully suppressed, causing \survey\ to fail at essentially the same 
density as for 0-hidden formulas, close to the satisfiability threshold, 
even though our $q$-hidden formulas continue to be satisfiable at all densities.  
In contrast, the 2-hidden formulas of~\cite{2hidden} are solved by \survey\ 
up to a somewhat higher density $r \approx 4.8$.  
Thus it seems that the reweighting approach of $q$-hidden formulas 
does a better job of confusing \survey\ than hiding two complementary assignments  does.

For $q < q^*$, \survey\ succeeds up to somewhat higher densities, each of which 
matches quite closely the value $r_c(q)$ at which \zchaff's running time peaks 
(see Figure~\ref{fig:rc} below). 
Building on our conjecture that this is the density above which the only solutions 
are those close to the hidden assignment, we guess that \survey\ succeeds 
for $r < r_c(q)$ precisely because the local gradient in the density of solutions 
pushes it towards the ``alternate'' solutions with $\alpha < 1/2$.  Above $r_c(q)$, 
these solutions no longer exist, and \survey\ fails because the clauses 
send deceptive messages, demanding that variables be set opposite to the 
hidden assignment.

\subsection{\walksat}

\begin{figure}[ht]
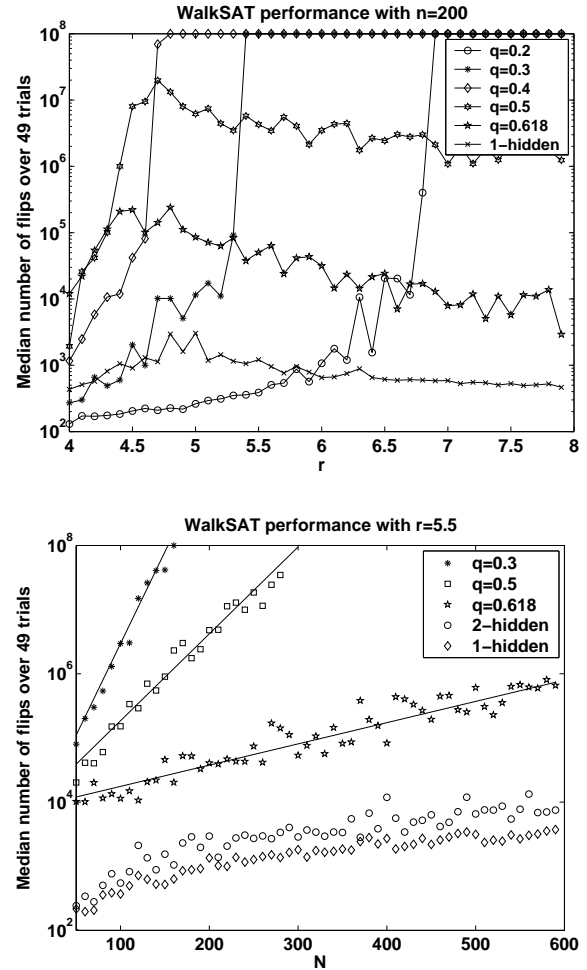

\begin{center}
\begin{minipage}{3in}
\begin{center}
        \centerline{\hbox{
        \psfig{figure=walkr.ps,width=3in}
        }}
\end{center}
\end{minipage}\    \
\begin{minipage}{3in}
\begin{center}
        \centerline{\hbox{
        \psfig{figure=walkn.ps,width=3in}
        }}
\end{center}
\end{minipage}
\end{center}
\vspace*{-.6cm}
\caption{The upper part shows \walksat's median running time over 49 trials on
$q$-hidden formulas with $n=200$ and $r$ ranging from $4$ to $8$; 
the lower part shows the median running time with $r=5.5$ and $n$ ranging from $50$ to $600$.}
\label{fig:walk}
\end{figure}

We conclude with a local search algorithm, \walksat. 
For each formula, we did up to $10^4$ restarts, with 
$10^4$ steps per attempt, 
where each step does a random or greedy flip with equal probability.
In the upper part of~Figure~\ref{fig:walk} we measure
\walksat's performance on $q$-hidden formulas with a range of values of 
$q$, including $q=1$, $q=q^*$, and deceptive values of $q$ ranging 
from $0.2$ to $0.5$.  We used $n=200$ and let $r$ range from $4$ to $8$.  
Even for these relatively small formulas, we see that for the three most 
deceptive values of $q$, there is a density at which the median running time 
jumps to $10^8$, indicating that \walksat\ fails to solve these formulas.  
For instance, $q$-hidden formulas with $q=0.4$ appear to be unfeasible 
for \walksat\ for, say, $r > 5$.

We believe that, consistent with the discussion above, local search algorithms like \walksat\ 
greedily follow the gradient in the density of solutions $f(\alpha)$.   For $q < q^*$, 
this gradient is deceptive, and lures \walksat\ away from the hidden assignment.  
At densities below $r_c(q)$, there are many alternate solutions with $\alpha < 1/2$ 
and \walksat\ finds one of them very easily; but for densities above $r_c(q)$, 
the only solutions are those near the hidden assignment, and \walksat's 
greed causes it to wander for an exponentially long time in the wrong region.  
This picture is supported by the fact that, as Figure~\ref{fig:rc} shows below, the density 
at which \walksat's running time jumps upward closely matches the thresholds $r_c(q)$ 
that we observed for \zchaff\ and \survey.  

The lower part of Figure~\ref{fig:walk} looks at \walksat's median running time 
at a fixed density as a function of $n$.  We compare 1-hidden and 2-hidden formulas
with $q$-hidden ones with $q=q^*$ and two deceptive values, $0.5$ and $0.3$.  
We choose $r=5.5$, which is above $r_c(q)$ for all three values of $q$.  
The running time of 1-hidden and 2-hidden formulas is only polynomial~\cite{2hidden,Barthel}.  
In contrast, even in the balanced case $q=q^*$, the running time is 
exponential, and the slope of this exponential increases dramatically as we decrease $q$ 
and make the formulas more deceptive.  We note that it might be possible to develop a 
heuristic analysis of \walksat's running time in the deceptive case using the methods of~\cite{monassonwalksat,monassonbook}.

\section{The threshold density}

\begin{figure}[ht]
\centerline{\hbox{
        \psfig{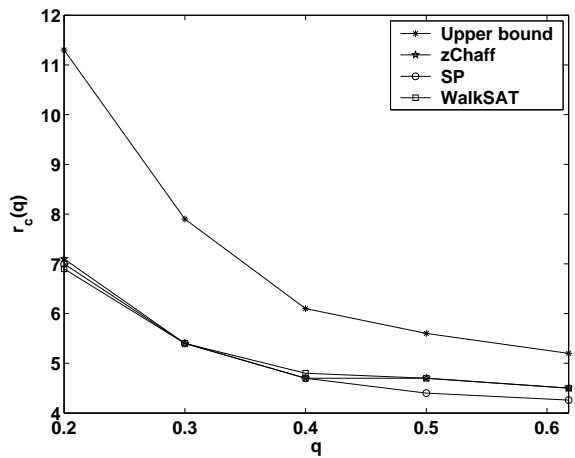}
}}
\caption{The density $r_c(q)$ at which the running time of \zchaff\ peaks, \walksat\ peaks or exceeds $10^8$ flips, and \survey\ stops working.  We conjecture all of these events occur because at this density the alternate solutions with $\alpha < 1/2$ disappear, leaving only those close to the hidden assignment.  Shown also is the analytic upper bound described in the text.}
\label{fig:rc}
\end{figure}

As we have seen, there appears to be a characteristic density $r_c(q)$ for each value of $q \le q^*$ 
at which the running time of DPLL algorithms like \zchaff\ peaks, at which \walksat's running time
becomes exponential, and at which \survey\ ceases to work.  We conjecture that in all three cases, 
the key phenomenon at this density is that the solutions with $\alpha < 1/2$ disappear, leaving 
only those close to the hidden assignment.  Figure~\ref{fig:rc} shows our measured values of 
$r_c(q)$, and indeed they are quite close for the three algorithms.  We also show the 
analytic upper bound on $r_c(q)$ resulting from setting $\max \{ f(\alpha) \mid \alpha \le 1/2 \} = 1$, 
above which the expected number of solutions with $\alpha \le 1/2$ is exponentially small.

\remove{
\begin{table}\label{tab:survey}
\begin{center}
\vspace*{0.3cm} \hspace*{0.0cm} $
\begin{array}{l|ccccc}
 \hspace*{2.7cm} q      & \makebox[0cm]{$q^*$}  & \makebox[0cm]{0.5}         &  \makebox[0cm]{0.4}        &  \makebox[0cm]{0.3}       &   \makebox[0cm]{0.2}  \\   \hline
 \mbox{$r_c(q)$ from calculation}     & 5.2 & 5.6 & 6.1    & 7.9 & 11.3  \\
 \mbox{$r_c(q)$ from \zchaff}     & 4.5 & 4.7 & 4.7     & 5.4 & 7.1  \\
\mbox{$r_c(q)$ from \survey}    & 4.26 & 4.4 & 4.7     & 5.4 & 7.0  \\
\mbox{$r_c(q)$ from \walksat}    & 4.5 & 4.7 & 4.8     & 5.4 & 6.9  \\
\end{array}
$ \vspace*{0.2cm}
\caption{$r_c(q)$ observed from calculation, \zchaff, \survey\ and \walksat}
\end{center}
\end{table}
}

\section{Conclusions}

We have introduced a simple new way to hide solutions in 3-SAT problems 
that produces instances that are both hard and satisfiable.  
Unlike the 2-hidden formulas of~\cite{2hidden} where the attraction of the hidden assignment
is cancelled by also hiding its complement, here we eliminate this attraction 
by reweighting the distribution of clauses as in~\cite{achperes}.  
Indeed, by going beyond the value of the parameter $q$ that makes 
our $q$-hidden formulas balanced, we can create {\em deceptive} formulas
that lead algorithms in the wrong direction.  Experimentally, our formulas 
are as hard or harder for DPLL algorithms as 0-hidden formulas, i.e., 
random 3-SAT formulas without a hidden assignment; 
for local search algorithms like \walksat, 
they are much harder than 0-hidden or 2-hidden formulas, taking 
exponential rather than polynomial time.  Our formulas are also
amenable to all the mathematical tools developed for the study of random 3-SAT; 
here we have calculated their expected density of solutions as a function 
of distance from the hidden assignment, and used the method of differential 
equations to show that \uc\ fails for them at the same density as it does for 
0-hidden formulas.  

We close with several exciting directions for future work:
\begin{enumerate}
\item Confirm that there is a single threshold density $r_c(q)$ at which 
a) the alternate solutions far from the hidden assignment disappear, 
b) the running time of DPLL algorithms is maximized, 
c) \survey\ stops working, and 
d) the running time of \walksat\ becomes exponential.
\item Prove that simple DPLL algorithms take exponential time
for $r > r_c(q)$, in expectation or with high probability.
\item Calculate the variance of the number of solutions as a function of $\alpha$, 
and giving improved upper and lower bounds on the distribution of solutions and $r_c(q)$.
\end{enumerate}

\section*{Acknowledgments}

H.J. is supported by an NSF Graduate Fellowship.
C.M. and D.S. are supported by NSF grants CCR-0220070, EIA-0218563, and PHY-0200909.  
C.M. thanks Tracy Conrad and She Who Is Not Yet Named for their support.

\newcommand{\proc}{}
\bibliography{qhiddenaaai}

\begin{thebibliography}{}

\bibitem[\protect\citeauthoryear{Achlioptas \& Moore}{2002a}]{achmoorecoloring}
Achlioptas, D., and Moore, C.
\newblock 2002a.
\newblock Almost all graphs with average degree 4 are 3-colorable.
\newblock {\em STOC}  199--208.

\bibitem[\protect\citeauthoryear{Achlioptas \& Moore}{2002b}]{achmooreksat}
Achlioptas, D., and Moore, C.
\newblock 2002b.
\newblock The asymptotic order of the random $k$-{SAT} threshold.
\newblock {\em FOCS}  779--788.

\bibitem[\protect\citeauthoryear{Achlioptas \& Peres}{2003}]{achperes}
Achlioptas, D., and Peres, Y.
\newblock 2003.
\newblock The threshold for random $k$-{SAT} is $2^k (\ln 2 - o(k))$.
\newblock {\em STOC}  223--231.

\bibitem[\protect\citeauthoryear{Achlioptas, Beame, \& Molloy}{2001}]{abm}
Achlioptas, D.; Beame, P.; and Molloy, M.
\newblock 2001.
\newblock A sharp threshold in proof complexity.
\newblock {\em STOC}  337--346.

\bibitem[\protect\citeauthoryear{Achlioptas \bgroup \em et al.\egroup
  }{2000}]{quasigroup2}
Achlioptas, D.; Gomes, C.; Kautz, H.; and Selman, B.
\newblock 2000.
\newblock Generating satisfiable problem instances.
\newblock {\em AAAI}  256.

\bibitem[\protect\citeauthoryear{Achlioptas, Jia, \& Moore}{2004}]{2hidden}
Achlioptas, D.; Jia, H.; and Moore, C.
\newblock 2004.
\newblock Hiding satisfying assignments: two are better than one.
\newblock {\em AAAI}  131--136.

\bibitem[\protect\citeauthoryear{Asahiro, Iwama, \& Miyano}{1996}]{Asahiro}
Asahiro, Y.; Iwama, K.; and Miyano, E.
\newblock 1996.
\newblock Random generation of test instances with controlled attributes.
\newblock {\em DIMACS Series in Disc. Math. and Theor. Comp. Sci.} 26.

\bibitem[\protect\citeauthoryear{Barthel \bgroup \em et al.\egroup
  }{2002}]{Barthel}
Barthel, W.; Hartmann, A.; Leone, M.; Ricci-Tersenghi, F.; Weigt, M.; and
  Zecchina, R.
\newblock 2002.
\newblock Hiding solutions in random satisfiability problems: A statistical
  mechanics approach.
\newblock {\em Phys. Rev. Lett.} 88(188701).

\bibitem[\protect\citeauthoryear{Chao \& Franco}{1986}]{chaofranco}
Chao, M., and Franco, J.
\newblock 1986.
\newblock Probabilistic analysis of two heuristics for the 3-satisfiability
  problem.
\newblock {\em SIAM J. Comput.} 15(4):1106--1118.

\bibitem[\protect\citeauthoryear{Cheeseman, Kanefsky, \&
  Taylor}{1991}]{cheeseman}
Cheeseman, P.; Kanefsky, R.; and Taylor, W.
\newblock 1991.
\newblock Where the really hard problems are.
\newblock {\em IJCAI}  163--169.

\bibitem[\protect\citeauthoryear{Cocco \& Monasson}{2004}]{monassontcs}
Cocco, S., and Monasson, R.
\newblock 2004.
\newblock Heuristic average-case analysis of the backtrack resolution of random
  3-satisfiability instances.
\newblock {\em Theor. Comp. Sci.} 320:345--.

\bibitem[\protect\citeauthoryear{Cocco \bgroup \em et al.\egroup
  }{2005}]{monassonbook}
Cocco, S.; Monasson, R.; Montanari, A.; and Semerjian, G.
\newblock 2005.
\newblock Approximate analysis of search algorithms with ``physical'' methods.
\newblock In Percus, A.; Istrate, G.; and Moore, C., eds., {\em Computational
  Complexity and Statistical Physics}. Oxford University Press.

\bibitem[\protect\citeauthoryear{Hogg, Huberman, \& Williams}{1996}]{hh}
Hogg, T.; Huberman, B.; and Williams, C.
\newblock 1996.
\newblock Phase transitions and complexity.
\newblock {\em Artificial Intelligence} 81.

\bibitem[\protect\citeauthoryear{Johnson \bgroup \em et al.\egroup
  }{1989}]{Johnson}
Johnson, D.; Aragon, C.; McGeoch, L.; and Shevon, C.
\newblock 1989.
\newblock Optimization by simulated annealing: an experimental evaluation.
\newblock {\em Operations Research} 37(6):865--892.

\bibitem[\protect\citeauthoryear{Kautz \bgroup \em et al.\egroup
  }{2001}]{quasigroup1}
Kautz, H.; Ruan, Y.; Achlioptas, D.; Gomes, C.; Selman, B.; and Stickel, Ê.
\newblock 2001.
\newblock Balance and filtering in structured satisfiable problems.
\newblock {\em IJCAI}  351--358.

\bibitem[\protect\citeauthoryear{Kullmann}{2002}]{oksolver}
Kullmann, O.
\newblock 2002.
\newblock Investigating the behaviour of a {SAT} solver on random formulas.
\newblock Technical Report CSR 23-2002, University of Wales Swansea.

\bibitem[\protect\citeauthoryear{M\'{e}zard \& Zecchina}{2002}]{mz}
M\'{e}zard, M., and Zecchina, R.
\newblock 2002.
\newblock Random k-satisfiability: from an analytic solution to a new efficient
  algorithm.
\newblock {\em Phys. Rev. E} 66:056126.

\bibitem[\protect\citeauthoryear{Mitchell, Selman, \& Levesque}{1992}]{msl}
Mitchell, D.; Selman, B.; and Levesque, H.
\newblock 1992.
\newblock Hard and easy distributions of {SAT} problems.
\newblock {\em AAAI}  459--465.

\bibitem[\protect\citeauthoryear{Morris}{1993}]{Morris}
Morris, P.
\newblock 1993.
\newblock The breakout method for escaping from local minima.
\newblock {\em AAAI}  40--45.

\bibitem[\protect\citeauthoryear{Selman, Kautz, \& Cohen}{1996}]{Selman96}
Selman, B.; Kautz, H.; and Cohen, B.
\newblock 1996.
\newblock Local search strategies for satisfiability testing.
\newblock {\em 2nd DIMACS Challange on Cliques, Coloring, and Satisfiability}.

\bibitem[\protect\citeauthoryear{Semerjian \& Monasson}{2004}]{monassonwalksat}
Semerjian, G., and Monasson, R.
\newblock 2004.
\newblock A study of pure random walk on random satisfiability problems with
  ``physical'' methods.
\newblock {\em LNCS} 2919:120--.

\bibitem[\protect\citeauthoryear{Shaw, Stergiou, \& Walsh}{1998}]{ssw}
Shaw, P.; Stergiou, K.; and Walsh, T.
\newblock 1998.
\newblock Arc consistency and quasigroup completion.
\newblock {\em ECAI workshop on non-binary constraints}.

\bibitem[\protect\citeauthoryear{Van~Gelder}{1993}]{VanGelder}
Van~Gelder, A.
\newblock 1993.
\newblock Problem generator {\tt mkcnf.c}.
\newblock {\em DIMACS challenge archive}.

\bibitem[\protect\citeauthoryear{Zhang}{}]{zchaffsite}
Zhang, L.
\newblock {zChaff}.
\newblock ee.princeton.edu/\~{}chaff/zchaff.php.

\end{thebibliography}
\bibliographystyle{aaai}

\end{document}